# Speech Recognition for Endangered and Extinct Samoyedic languages

**Niko Partanen**
University of Helsinki
Finland
`niko.partanen@helsinki.fi`

**Mika Hämäläinen**
University of Helsinki
and Rootroo Ltd
Finland
`mika@rootroo.com`

**Tiina Klooster**
Luua Forestry School
Estonia
`tiinaklooster@gmail.com`

## Abstract

Our study presents a series of experiments on speech recognition with endangered and extinct Samoyedic languages, spoken in Northern and Southern Siberia. To best of our knowledge, this is the first time a functional ASR system is built for an extinct language. We achieve with Kamas language a Label Error Rate of 15%, and conclude through careful error analysis that this quality is already very useful as a starting point for refined human transcriptions. Our results with related Nganasan language are more modest, with best model having the error rate of 33%. We show, however, through experiments where Kamas training data is enlarged incrementally, that Nganasan results are in line with what is expected under low-resource circumstances of the language. Based on this, we provide recommendations for scenarios in which further language documentation or archive processing activities could benefit from modern ASR technology. All training data and processing scripts haven been published on Zenodo with clear licences to ensure further work in this important topic.

## 1 Introduction

Samoyedic languages are spoken in the Western Siberia, and Tundra Nenets also extends far to the European part of Northern Russia. These languages belong to the Uralic language family. All currently spoken Samoyedic languages are endangered (see Moseley (2010). Tundra Nenets is the largest language in the family, while some, such as Kamas, are extinct. Even extinct Samoyedic languages have, however, been documented to various degrees.

Documentation of Samoyedic languages has reached a mature stage in last decades. Three languages in this group have a recent monograph long grammars in English. These are Tundra Nenets (Nikolaeva, 2014), Forest Enets (Siegl, 2013) and Nganasan (Wagner-Nagy, 2018). Similarly resources on these languages have been steadily becoming available for researchers, often in connection with major language documentation projects, such as 'The word stock of the Selkup language as the main source of cultural and historical information about a moribund endangered native ethnicum of Western Siberia'[1], 'Enets and Forest Nenets'[2], 'Corpus of Nganasan Folklore Texts'[3], 'Documentation of Enets: digitization and analysis of legacy field materials and fieldwork with last speakers'[4], 'Tundra Nenets Texts'[5], 'Comprehensive Documentation and Analysis of Two Endangered Siberian Languages: Eastern Khanty and Southern Selkup'[6], 'Selkup Language Corpus (SLC)' (Budzisch et al., 2019), 'Nganasan Spoken Language Corpus' (Brykina et al., 2018), 'INEL Selkup Corpus' (Brykina et al., 2020), 'INEL Kamas corpus' (Gusev et al., 2019). Also Online Dictionary for Uralic Languages (Rueter and Hämäläinen, 2017) includes Tundra Nenets lexical material.

---

[1] `https://cordis.europa.eu/project/id/INTAS2005-1000006-8411`
[2] `https://dobes.mpi.nl/projects/nenets/`
[3] `https://iling-ran.ru/gusev/Nganasan/texts/`
[4] `https://elar.soas.ac.uk/Collection/MPI950079`
[5] `https://elar.soas.ac.uk/Collection/MPI120925`
[6] `https://elar.soas.ac.uk/Collection/MPI43298`

Our study here focuses on two of these languages: Nganasan and Kamas. Our topic is ASR, but we anchor this work into a wider context of computational resources and language documentation. The work has two goals: to examine feasibility of developing an ASR system for an extinct language, in the case of Kamas, and to investigate the usability of such a system in a real on-going endangered language documentation scenario that is presented by Nganasan. These scenarios are not wide apart from one another. Worldwide extinction of linguistic diversity has been recognized for the last 30 years (Krauss, 1992), and many languages are in a very endangered situation.

The models trained in this paper have been released and made openly accessible on Zenodo with a permanent DOI[7]. As both corpora used in our study are distributed with restrictive non-commercial license (CC-BY-NC-SA), we have also published our training and testing materials with the same license. We think open practices such as these will be gaining importance in the future, and we want to contribute to this development as well by our example (Garellek et al., 2020).

## 2 Context and Related Work

Despite the wide documentation activities, we have not witnessed large improvements in language technology and computational linguistics on these languages. Thereby one of the goals in this paper is to encourage further work on these languages, also through our published new datasets. There is a rule based morphological analyser for Nganasan (Endrédy et al., 2010), which, however, appears to be available only through a web interface, and is not open access.

What it comes to other Samoyedic languages, a rule based Tundra Nenets morphological analyser exists in the GiellaLT infrastructure (Moshagen et al., 2014)[8] with availability through UralicNLP (Hämäläinen, 2019). There are also early Selkup[9] and Nganasan[10] analysers in the same infrastructure. Also OCR models have been developed to target early writing systems on these languages (Partanen and Rießler, 2019b), with associated data package (Partanen and Rießler, 2018). This responds well to OCR challenges identified earlier for these languages by Partanen (2017). The vast majority of these languages have virtually no language technology at the moment, but as there are increasingly larger and larger corpora, the possibilities for future work are many.

One challenge in working with these endangered languages is that very few researchers are able to transcribe them confidently and accurately. In the past few years, however, speech recognition in endangered language context has seen significant improvements, especially in scenarios where there is only one single speaker. Adams et. al. (2018) report a reasonable accuracy under these circumstances already with just a few hours of transcribed data, with rapid increase in accuracy when there is more training data. They also present a comparison of models trained on different amounts of training data using Na and Chatino data, which also inspired our own comparative experiments.

We have also seen very recently large improvements in such systems on related Uralic languages, for example Zyrian Komi (Hjortnaes et al., 2020b; Hjortnaes et al., 2020a). We have also seen experiments where ASR is being integrated to the language documentation workflows, for example, in Papuan context (Zahrer et al., 2020). Most widely applied speech recognition systems have been Persephone (Adams et al., 2018), Elpis (Foley et al., 2018) and DeepSpeech (Hannun et al., 2014). In this paper, we present and discuss several experiments we have done using Persephone system.

## 3 Languages and Data

Nganasan is an endangered Samoyedic language spoken by Nganasans, a small ethnic group in Taimyr Peninsula, Northern Siberia (Janhunen and Gruzdeva, 2020). According to official statistics there are 470 Nganasans, from who approximately 125 speak the Nganasan language (Wagner-Nagy, 2018, 3,17). Despite languages endangerment, plenty of documentation work has been conducted (Leisio, 2006; Wagner-Nagy, 2014; Kaheinen, 2020). Largest available Nganasan corpus was published in 2018 (Brykina et al., 2018), and it was used in our study.

---

[7] https://zenodo.org/record/4029494
[8] https://github.com/giellalt/lang-yrk
[9] https://github.com/giellalt/lang-sel
[10] https://github.com/giellalt/lang-nio

Kamas is another Samoyedic language, representing the southern group of this branch of Uralic languages. Kamas was spoken in the slopes of Sayan mountains in the Central Siberia. It is believed that by the 19th century Kamas tribe consisted of only 130 people (Matveev, 1964). Kamas were forced to abandon their nomadic lifestyle in the beginning of 20th century, which, in connection with large societal changes, increased the contact with Russian speakers and led to a cultural assimilation (Klooster, 2015, 9). The last Kamas speaker was Klavdiya Plotnikova, who was born in 1895 in a small village of Abalakovo in Central Siberia. She worked with several linguists since 1960s, and this results in a sizable collection of Kamas recordings that are available in various archives.

In 2019, a corpus containing transcribed versions of these materials was published (Gusev et al., 2019). We used Klavdiya Plotnikova's part of the corpus in our Kamas experiments, as she contributes the vast majority of all Kamas materials that exists. In the Nganasan experiments, we used the data from three prominent speakers in the Nganasan Spoken Language Corpus, who are also mentioned in the Nganasan grammar based largely to the same materials (Wagner-Nagy, 2018, 30).

One of the most important preprocessing steps was to exclude from training all sentences that are longer than 10 seconds. This is a condition set by Persephone system, and a convention followed also in other investigations (Wisniewski et al., 2020, 30). Similarly, Hjortnaes et al (2020b) filtered Zyrian Komi corpus by this limit. This choice leaves open an obvious possibility to improve the current results. As the filtered portion of the corpus is relatively large, either finding a way to include longest segments into the training process, or splitting those into smaller units, would easily increase the amount of training data.

Preprocessing conventions were very similar with both corpora, although independent inspection of particularities of individual datasets was done. It is customary that we work with speech corpora includes an intensive preprocessing step. With the case of Kamas the work was greatly aided by having a specialist of Kamas in our team. In the case of Nganasan we worked primarily with the light shed by the project documentation, which was also an useful and realistic scenario.

As Nganasan corpus was significantly larger than Kamas, also more preprocessing was needed, probably reflecting the longer time frame where it has been created. We excluded all segments that were shorted than 400 milliseconds, and removed all empty annotations or annotations that contained only punctuation characters. There were several invisible Unicode space characters that were removed. Also all annotations that contained number written as number characters were excluded. We also removed from Nganasan corpus all instances of utterances that contained Cyrillic characters.

For both corpora the annotations that contained unclear words marked with HIAT conventions (Ehlich and Rehbein, 1976) were removed. When the transcriptions contained annotations for non-verbal expressions, including coughing or laughter, we chose to remove those extra annotations, but keep those transcriptions in the training data. Self-corrections were kept, but the hyphens and brackets around them were removed.

It has been asked previously to what degree the preprocessing of language documentation data can be automatized (Wisniewski et al., 2020). Based to our experience with these corpora we can say that a good deal of manual inspection and examination is necessary to understand how the raw data has to be processed to make it usable in ASR training. In our experience the actual transformation of corpus XML files into the structure expected by Persephone was relatively easy. Much more time was consumed by analysing the annotation conventions used in the corpus, and processing some of the mistakes in transcriptions. To this vein we can strongly recommend for different projects the approach suggested by (Partanen and Riessler, 2019a), where a team working with endangered languages of Barents region have integrated automatic testing and validation very deeply into their corpus workflows, thus ensuring the systematic and orderly presentation of the corpus.

## 4 Method and Experiment Design

Our model is a bi-directional long short-term memory (LSTM) (Hochreiter and Schmidhuber, 1997; Schuster and Paliwal, 1997), which is trained to pre-dict the character sequences from audio input. The loss function used in the training is a connectionist temporal classification (CTC) in order to make it possible to train the model with only a coarse alignment between the audio and text (Graves et al., 2006). As suggested by (Wisniewski et al., 2020) and (Adams et al., 2018), we use 3 hidden layers with 250 hidden units. Using the same settings as other experiment have done maximises the comparative value of current work. We train our models by using Persephone framework (Adams et al., 2018).

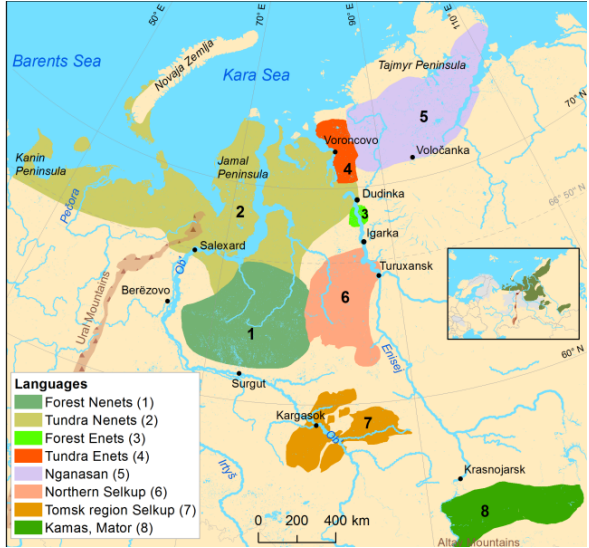

Figure 1: Distribution of the Samoyedic languages in the beginning of the 20 century (Timo Rantanen, BEDLAN)

### 4.1 Nganasan Tests

Although Nganasan corpus is fairly large, 28 hours according to corpus description, there are few individual speakers who are most prominent in the dataset. We selected all data from three such speakers, and trained individual models for all of them. To our knowledge experiments with Persephone system have not been carried out with tens of hours of data, and as earlier experiments have also focused to single speaker settings, we decided to continue along these lines. As the Nganasan speakers represent different amounts of transcribed data, the differences in accuracy can still give important information about this particularly low-resource setting.

### 4.2 Kamas Tests

After our preprocessing methods the Kamas corpus contains approximately 5 hours of utterances from a single speaker, above mentioned Klavdiya Plotnikova. Thereby the most obvious experiment to conduct is to train the model with all the speech we have from her, taking the original transcription as it is.

However, this also allows more varied experiments. One of the questions that are unclear in low-resource ASR is what to do with the word boundaries. In phoneme-level recognition usually practices with Persephone they often are omitted. Also in experiments done with other tools, such as DeepSpeech, specific language models have been used to insert spaces into correct places (Hjortnaes et al., 2020b). The second experiment with Kamas arises from this starting point: let's just leave the word boundaries as predicted labels.

One problem with the word boundaries must be their nature as a higher level construct. In real speech they often do not appear as pauses. As forced alignment tools have already developed relatively far, we opted for the MAUS system (Kisler et al., 2012) to align our data automatically. More specifically, we used the functionality provided in the emuR R package (Winkelmann et al., 2017). The language was left unspecified, and the alignment used grapheme to phoneme -mapping for the original transcriptions, and returned phoneme aligned SAMPA and IPA versions (Reichel, 2012; Reichel and Kisler, 2014). The mapping will be published openly with the other resources described here. It must be noted that there are minor differences between the original transcription and IPA representation. These are primarily about the how the long vowels are presented, as the original transcript was primarily split to characters, whereas the converted IPA has more phonemic units as individual labels.

This process gave us two more transcriptions versions: Plain IPA text, and an IPA version where only those word boundaries were retained which occurred within natural pauses. This work was highly experimental, and we did not correct the segmentations manually. Same Kamas data will be included in manually corrected DoReCo corpus (Paschen et al., 2020), which will allow better inspection of these features. Our primary goal in this experiment was simply to investigate whether essentially very minor changes in transcriptions impact the result, and to see if different representation of word boundaries brings any benefits.

| Experiment | Utterances | Minutes | LER |
|---|---|---|---|
| 1 | 1152 | 108 | 0.334 |
| 2 | 512 | 57 | 0.930 |
| 3 | 704 | 43 | 0.892 |

Table 1: Nganasan experiments with three different speakers.

| Exp. | Description | LER |
|---|---|---|
| 4 | Original transcript, no spaces | 0.226 |
| 5 | Original transcript, with spaces | 0.195 |
| 6 | IPA transcript, no spaces | 0.149 |
| 7 | IPA transcript, real pauses | 0.243 |

Table 2: Kamas experiments with 4224 training samples, 266 minutes.

### 4.3 Gradual Data Augmentation Test

In order to evaluate the importance of the amount of transcribed minutes and hours, we designed additional tests. In these experiments we use the exactly same Kamas corpus as in the Experiment 6, but take only smaller portions of it, that are augmented gradually. As the maximum amount of data was close to 5 hours, we selected intervals that should represent realistically increasing corpus size, and thereby show where the most important thresholds lie.

These experiments are described in Table 3. While discussing the results we have also compared our error rates to those reported in other studies, in order to understand better how the variation we see connects to earlier studies with different languages.

## 5 Results

In this section, we present the results of the different models. These results are reported as a LER (label error rate) score. In practice, this is a measurement similar to CER (character error rate) that is widely used in studies focusing on text normalization and OCR correction (see (Tang et al., 2018; Veliz et al., 2019; Hämäläinen and Hengchen, 2019)).

### 5.1 Nganasan Results

In Nganasan experiment we selected utterances from three most prominent speakers. Table 1 shows the amount of data that we used and the accuracy reached.

We can easily conclude that the results were not successful in all experiments. In the cases where we had less than an hour of transcriptions, the quality was extremely low. When the label error rate is this high the model does not produce a useful result. However, there was a clear improvement with one speaker for who we had more training material. Brief example and discussion is provided in Section 6.

### 5.2 Kamas Results

Compared to the Nganasan experiment, the Kamas results are very different. Indeed, the results we achieve are very high, and on par with the best scores reported elsewhere for Persephone. We argue that the primary reason to this is sufficient amount of training data. Table 2 shows these results in detail.

In Experiment 4 we trained Persephone on the original Kamas transcriptions, without word boundaries separately marked, and with no modifications to the existing transcriptions. In the Experiment 5 the space characters were left to their original places. Surprisingly the result is significantly better with the word boundaries than without them.

Since Experiments 4 and 6 use extremely similar training data, just in different transcription system, we would had assumed the results to be very similar. We see, however, very large difference between the models. As we did not run the experiments multiple times, it is left open whether the difference can be caused from different random seeds. In our error analysis some possible reasons for these differences are discussed further.

The Experiment 5, however, was necessary in order to evaluate with more confidence the results of Experiment 7. Between these experiments the only difference was in detected pauses, instead of original word boundaries. The procedure was described in Section 4. As the Experiment 7 produces the worst results, we must conclude that this experiment was not successful. However, since the presence of word boundaries as their own tokens has small impact to the accuracy, and as they are useful information, this model may still be favoured in actual use.

| Experiment | Utterances | Minutes | LER |
|---|---|---|---|
| 6-1 | 448 | 28 | 0.612 |
| 6-2 | 896 | 57 | 0.254 |
| 6-3 | 1856 | 117 | 0.224 |
| 6-4 | 2816 | 177 | 0.176 |
| 6-5 | 3776 | 238 | 0.190 |

Table 3: Gradual data augmentation experiments

All Kamas models are relatively good, and the accuracy is inspected closer in Section 6. We see clearly in the Figure 2 that although there are minor differences, when the model has sufficient amount of training data the accuracy does not significantly change. We also cannot entirely exclude the possible impact of random run-time differences when the results are very close to one another.

### 5.3 Gradual Data Augmentation Test Result

The goal of this experiment was to investigate how the model's accuracy changes when the amount of training data is increased. In the past we have seen various tests with different corpora, often reaching very good results, as discussed in Section 2. The results of this experiment are presented in Table 3.

The major result we find here is that soon after containing two hours of training data, the models show extremely modest improvements. The largest improvement takes place between half an hour and a full hour. Especially when we compare the results to those reported for different languages by Wisniwski et. al. (2020), it appears that the amount of training data is the main denominator that impacts the models accuracy. Na language model is essentially as good as Kamas model, which reaches it's maximal accuracy after three hours of training data. There are possible exceptions, for example, Duoxu model is relatively good compared to its small size. Even then, it fits to the general curve very well.

Based on this comparison, more training data is not necessarily better, and the benefits decrease after certain level has been reached. We essentially have repeated the results of Adams et al. (2018) on Na and Chatino. We will discuss this further in Section 7, but this clearly gives us some guidelines of how much transcriptions are needed at the moment to achieve the best possible accuracy. This also contextualizes the Nganasan results, and explains why one of the models was much better than the others.

## 6 Error Analysis

Our error analysis focuses mainly on Kamas, since with this language we achieved a very high level of accuracy. In our error analysis the numbered lines in examples correspond to experiments described and numbered in Section 4. We can, however, state briefly that two Nganasan models with worst accuracy predict mainly short character sequences, essentially repeating the same fixed string. This prediction is, naturally, not useful. With the best Nganasan model, however, the result could already be useful as preliminary transcription stage. We see in Example (1) that the errors are primarily connected to vowel length, and most of the words and morphemes start to be recognizable.

(1) Mənə bəbəəd'əətənɨ is'üðəm hüətə.
 mənəbəbəd'ətənis'ühuhətə
 'I will be all the time at the old place.'

Next examples are all from the Kamas corpus. Some of the mistakes different models make seem to be systematic. Examples (2) and (3) show that especially consonant sequences are challenging for the model. Both /ll/ or /tt/ come out systematically as single consonants.

(2) Ujabə ajirbi mĭnzərzittə.
 1: ujabajrbimĭnzərzitəo
 2: ujabaj irbi mĭnzərzitə
 3: ujabajirbimɨnzirzitə
 4: ujabajirbimɨnzərzitəo
 'He was reluctant to cook his meat.'

Especially in the second phoneme in Example (3) we see wide variation in the predicted vowel. This appears to be very common with reduced vowels.

(3) Mĭlleʔbi, mĭlleʔbi, ej kuʔpi.
 1: mĭleʔtimĭleʔtiejkuʔpiö
 2: müleʔpi mĭleʔtə ej kuʔpi
 3: nuleʔbəmɨleʔtəejkuʔpi
 4: mɨleʔpimɨleʔpiejkuʔpia
 'He went, he went, he did not kill.'

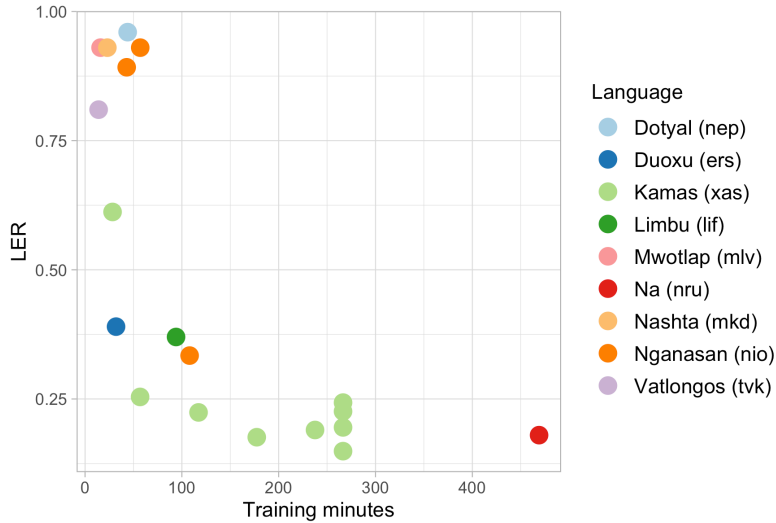

Figure 2: Results of our Nganasan and Kamas experiments compared with Wisniewski (2020)

In Example (4) we see how self-corrections have been treated in the original transcription and our various experiments. The model with original word boundaries is able to predict the spaces correctly, whereas the model with natural spaces only captures some of them. We see in this example also that none of the models recognize /b/ that is in the end of self correction. This plosive is not fully realized, which makes it acoustically very different from other instances of this phoneme. We can also notice that some of the models have a tendency to drop glottal stops, although some combinations show regularity.

(4) I dĭ poʔto (kuzab-) kuzazi mobi.
 1: idĭpoʔtkuzakuzaziʔmobi
 2: i dĭ poʔto kuza kuzaziʔ mobi
 3: idɪpoʔtokuzakuzazimobi
 4: idɪ potəkuzakuzaziʔ mobia
 'This goat became man.'

Although not intended for ASR error analysis, we decided to use OCR tool Calamari's error evaluation method (Wick et al., 2018). This gave us character level error information. Most frequent individual errors were related to letters /t/ and /l/. This seems to be related to them occurring as both short and long phonemes, and the models had particular problems to learn this distinction. Besides this we can see that many of the most prominent errors are related to vowels that share a very similar place of articulation or other properties: /e/ : /ə/, /o/ : /u/, /ɪ/ : /i/, /ø/ : /o/, /y/ : /u/, /ɐ/ : /a/. Within consonant system the errors between nasals /n/ : /m/ : /ŋ/ are frequent, and also glottal stop is often replaced with zero. We can also point that the error /t/ : /d/ is common, but other systematic errors related to voicing opposition cannot be found. These differences can be compared to the error analysis reported from Yongning language, where the confusions between characters also appeared to be related to acoustic realities (Michaud et al., 2020).

When our error analysis was repeated with the original transcription, which generally gave much worse results than IPA, very curious picture starts to emerge. Especially vowels containing diacritics were often misrecognized or omitted from prediction. This hints there is possibly something in this representation of the texts that does not pass correctly through the system and needs to be thoroughly investigated. Further research should be conducted with different character representations and refined data preprocessing. Also in-depth investigation of how the strings are passed to the ASR system internally could reveal more information about how, for example, different combining characters are treated. As we have published the training data and trained models openly, this examination can be continued easily.

# 7 Conclusion

We conducted several experiments on speech recognition of endangered and extinct languages. The most significant result is that we can identify a clear threshold of few hours after which the current models do not show clear decrease in the label error rate. We also show that differences in transcription system do not cause significant differences between models, although there is enough variation that ideal representation should be investigated. We do not see large differences in whether we use IPA or a project's internal transcription convention. As all these transcriptions are phonemic at fairly same level, the lack of differences is not surprising. Best results were achieved without word boundaries, but also the experiments with all word boundaries were encouraging enough that we would suggest testing a model trained with those intact. We aligned transcriptions to detect only word boundaries that correspond to natural pauses in speech, but the results of this were not better than in other experiments. We presume that the question here lies in the possibly weak quality of this automatic segmentation, and the experiment should be repeated with manually corrected version of the data.

As the transcription bottleneck is a major problem in linguistic fieldwork and documentation of endangered languages, our work sheds light to possible emerging working methods. This is also a question of resource allocation: when the language is rapidly disappearing, should we focus into recording more or into refinement and transcription of the existing materials? The situation is especially concerning when there are only a few individual elder speakers.

We hope our work offers new insight into this complicated question. Kamas has been extinct for more than 30 years, yet we are able to build a relatively good speech recognition model from just two hours of transcribed speech. This is a realistic amount, and not particularly much in the context where contemporary language documentation projects usually have budgets that cover several researchers work for multiple years.

Automatic transcription with ASR tools is a fast moving target. The results in few years will certainly be entirely different from what we are seeing now. Thereby making recommendations is also a complicated matter. However, we would argue, based to our results, that after one hour is transcribed for an individual speaker, training an ASR model to speed up the transcription work should already take place. In the same vein, this could be taken into account when working with endangered languages with very small speech communities. Recording and transcribing different speakers widely, with substantial transcription base for each speaker, seems to be the best way to take advantage from currently available ASR systems. We are clearly moving into a situation where manual transcription of everything is not the only option.

Future research should also focus into moving from single speaker systems into ASR that can work with multiple speakers, including unknown speakers. Very encouraging results were reported recently with only 10 minutes of training data, with the use of pretraining on unannotated audio and using a language model (Baevski et al., 2020). Since unannotated audio is available for virtually all language documentation projects, and also text corpora are becoming increasingly available and have proven useful (Hjortnaes et al., 2020a), there are certainly possibilities to experiment with these methods also in what it comes to language documentation context.

There is also evidence that systems other than Persephone could deliver better results, which is to be expected as the field evolves. Gupta and Boulianne (2020) reached a phoneme error rate of 8.7% on 3.1 hours of Cree training data, and their comparison of different systems showed significant improvement to other currently available methods, among those Persephone. This would suggest that there are possibilities to improve also from the Persephone results we have reported now.

The Persephone models that we have trained can be used with Cox's (2019) ELAN extension, or programmatically using Python. We have published both the models and the training datasets[11] in order to encourage further experiments on this important topic, and also to allow Nganasan researchers to benefit from our results. Although the majority of Kamas materials are already transcribed, we believe our results are relevant and valuable for the work being done with endangered and extinct languages.

---

[11]https://zenodo.org/record/4029494